\documentclass[runningheads]{llncs}
\usepackage{graphicx}
\usepackage{tikz}
\usepackage{comment}
\usepackage{amsmath,amssymb} % define this before the line numbering.
\usepackage{color}
\usepackage{booktabs}
\usepackage{threeparttable}
\usepackage{subfigure}
\usepackage{multirow}
\usepackage[colorlinks,
            linkcolor=red,
            anchorcolor=blue,
            citecolor=green]{hyperref}
\usepackage{floatrow}
\newfloatcommand{capbtabbox}{table}[][\FBwidth]

\begin{document}
% \linenumbers
\pagestyle{headings}
\mainmatter
\def\ECCVSubNumber{8}  % Insert your submission number here

\title{Color-$S^{4}L$: Self-supervised Semi-supervised Learning with Image Colorization} % Replace with your title

%\begin{comment}
%\titlerunning{ECCV-20 submission ID \ECCVSubNumber} 
%\authorrunning{ECCV-20 submission ID \ECCVSubNumber} 
%\author{Anonymous ECCV submission}
%\institute{Paper ID \ECCVSubNumber}
%\end{comment}
%******************

%\begin{comment}
\titlerunning{Color-$S^{4}L$: Semi-supervised Image Classification with Self-supervisions}
\author{Hanxiao Chen\inst{1}}
\authorrunning{Hanxiao Chen}
\institute{Department of Automation, Harbin Institute of Technology \\
\email{hanxiaochen@hit.edu.cn}}
%\end{comment}
%******************
\maketitle

\begin{abstract}
This work addresses the problem of semi-supervised image classification tasks with the integration of several effective self-supervised pretext tasks. Different from widely-used consistency regularization within semi-supervised learning, we explored a novel self-supervised semi-supervised learning framework (\textbf{Color-$S^{4}L$}) especially with image colorization proxy task and deeply evaluate performances of various network architectures in such special pipeline. Also, we demonstrated 
its effectiveness and optimal performance on CIFAR-10, SVHN and CIFAR-100 datasets in comparison to previous supervised and semi-supervised optimal methods. 
\end{abstract}

\section{Introduction}
Vision serves as the most promising way to explore machine learning instances within multiple benchmarks, such as image classification \cite{2}, action recognition \cite{1,51}, video segmentation \cite{24}. To tackle the fundamental weakness of supervised learning that demands for a vast amount of human-labeled data which is costly to collect and scale up, an emerging body of research on semi-supervised learning \cite{8}, few-shot learning \cite{4}, self-supervised learning \cite{2,24}, and transfer learning \cite{5} have dedicated towards learning paradigms that enable machines to recognize novel perception concepts by leveraging limited labels.

Within this effort, semi-supervised learning (SSL) matches our human learning patterns that conduct tasks well after developing compatible concepts by mastering some correct label information. Also, self-supervised visual representation learning has demonstrated huge potentials \cite{2,3,21,22} on tough computer vision tasks. In general, the aim of self-supervised learning is utilizing pretext tasks (i.e., Image Rotation \cite{26}, Geometric Transformation \cite{39} to learn intermediate semantic or structural features from large-scale unlabeled data and then transfer the representation to diversified downstream tasks like image classification \cite{2,21,22} and video segmentation \cite{24,25} effectively. 

Inspired by reasonable results of semi-supervised learning with self-supervised regularization \cite{23}, our work focuses on such novel but challenging topic, even explores the ingenious combination of various self-supervised surrogate tasks with the normal semi-supervised framework and designs an effective algorithm Color-$S^{4}L$ to tackle the semi-supervised image classifier issue. In general, Color-$S^{4}L$ belongs to a class of approaches \cite{6,7,15} that produce auxiliary labels from unlabeled data without manual annotations, which are utilized as credible targets along with labeled samples. The whole framework of Color-$S^{4}L$ is demonstrated in \textbf{Fig.1}. After conducting extensive experiments on Color-$S^{4}L$ with diverse network trunks like Convolutional Neural Networks (ConvNet), Wide Residual Network (WRN), etc, we find Color-$S^{4}L$ achieves competitive or state-of-the-art performance against previous SSL methods for both supervised and semi-supervised image classification with no additional hyper-parameters to tune.
\vspace{-5mm}
\begin{figure}
\centering
\includegraphics[width=12cm]{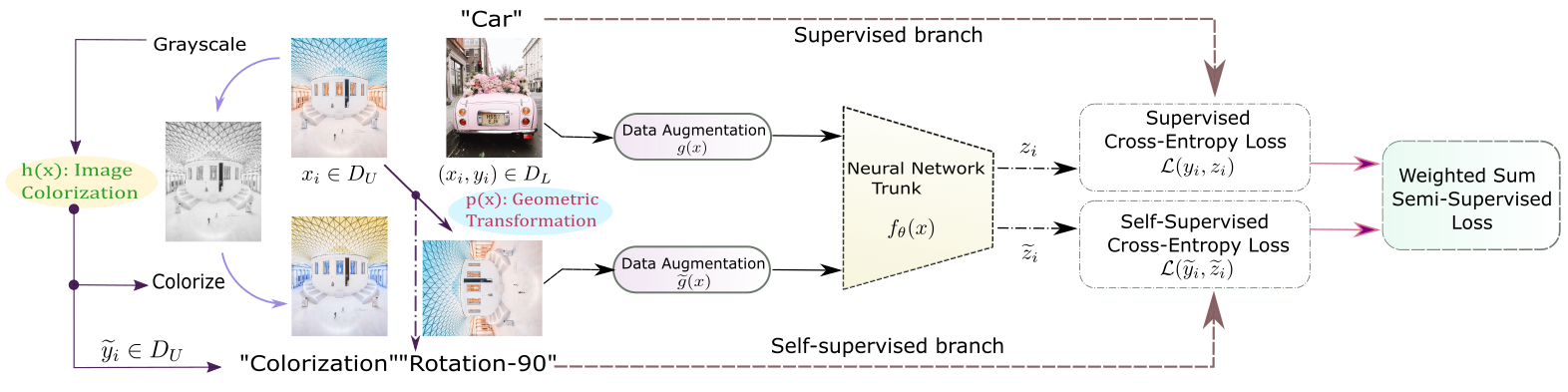}
\caption{Color-$S^{4}L$ architecture for semi-supervised image classification. The left of pipeline denotes two kinds of self-supervised proxy tasks which include \textbf{Image Colorization} and \textbf{Image Rotation}. Also, we employ \textbf{Geometric Transformation} function $p(x)$ to produce 6 proxy labels defined as image rotation in multiples of 90 degrees ($[0^{\circ},90^{\circ},180^{\circ},270^{\circ}]$) along with horizontal(left-right) and vertical(up-down) flips like \cite{23}. In addition, we especially utilize \textbf{Image Colorization} function $h(x)$ to create the $7^{th}$ auxiliary label to strengthen the existing self-supervisions on unlabeled data within semi-supervised learning paradigm.}
\label{1111}
\end{figure}
\vspace{-15mm}
\section{Methodology}
\vspace{-2mm}
\subsection{Overview of Color-$S^{4}L$ model} \vspace{-2mm}
As shown in \textbf{Fig.1}, we split the image data into two mini-batches of training sets with the same number of examples at each training step, one is the labeled input-target pairs $(x,y)\in D_{L}$ and another contains unlabeled inputs $x\in D_{U}$. The goal of Color-$S^{4}L$ is training a prediction function $f_{\theta}(x)$ which is parameterized by $\theta$, where the combination of $D_{L}$ and $D_{U}$ may benefit the model to achieve significantly better prediction performance than the single $D_{L}$. Therefore, we input two training samples separately in each supervised and self-supervised branch with a shared CNN backbone $f_{\theta}(x)$. Since there is a large amount of unlabeled data in the self-supervised line, labeled examples will represent a mini-batch repeatedly in Color-$S^{4}L$. Also, we operate forward-propagation on $f_{\theta}(x)$ both in the labeled branch $x_{i \in D_{L}}$ and the unlabeled one $x_{i\in D_{U}}$, resulting in softmax prediction vectors $z_{i}$ and $\tilde{z_{i}}$. Next we compute the Color-$S^{4}L$’s loss function both with the supervised cross-entropy loss $L_{super}(y_{i},z_{i})$ applying ground truth labels $y_{i}$ and the self-supervised cross-entropy loss $L_{self}(\tilde{y_{i}},\tilde{z_{i}})$ using proxy labels $\tilde{y_{i}}$ generated by multiple auxiliary tasks. Following \cite{23} the parameters $\theta$ can be learned via backpropagation techniques by minimizing the multi-task Color-$S^{4}L$ objective function defined as the weighted sum semi-supervised loss:
\footnote{After exploring different values of the parameter $\omega$ like 1,1.2,1.5,2, etc, we found $\omega$=1 yields consistent results across all datasets and CNN trunks similar to \cite{23}.}
\begin{align}
L_{Color-S^{4}L} = L_{super}(y_{i},z_{i})+\omega *L_{self}(\tilde{y_{i}},\tilde{z_{i}}) 
\end{align}
\ \ \ \ In addition, we need to apply multiple effective self-supervised pretext tasks to generate reasonable proxy labels independently in the self-supervised branch. Thus, we can match different tasks with different labels respectively. In detail, for \textbf{image rotation} we enable it to produce 4 labels such as $\tilde{y}=0,1,2,3$ since it serves as a 4-classification model to predict degrees corresponding to $[ 0^{\circ},90^{\circ},180^{\circ},270^{\circ}]$. Next, we choose the horizontal (left-right) and vertical (up-down) flips in \textbf{geometric transformation} to produce another two surrogate labels. To emphasize, we especially embed our self-trained \textbf{image colorization} model into the Color-$S^{4}L$ to recognize color changing within the self-supervised branch. Different from the techniques that produce transformation operations on images directly, we trained a new image colorization model with the extended Encoder-Decoder Network like \cite{35} (\textbf{Fig.2}) and transferred it on unlabeled data to yield another surrogate label (i.e. $\tilde{y}=7$) for the colorized samples. As for more details of training image colorization model and its experimental results, you are recommended to check our whole research report. \url{https://github.com/2000222/HIT-Computer-Vision/blob/master/ECCV_2020_HanxiaoChen.pdf}
\vspace{-8.2mm}
\begin{figure}
\centering
\includegraphics[width=12cm]{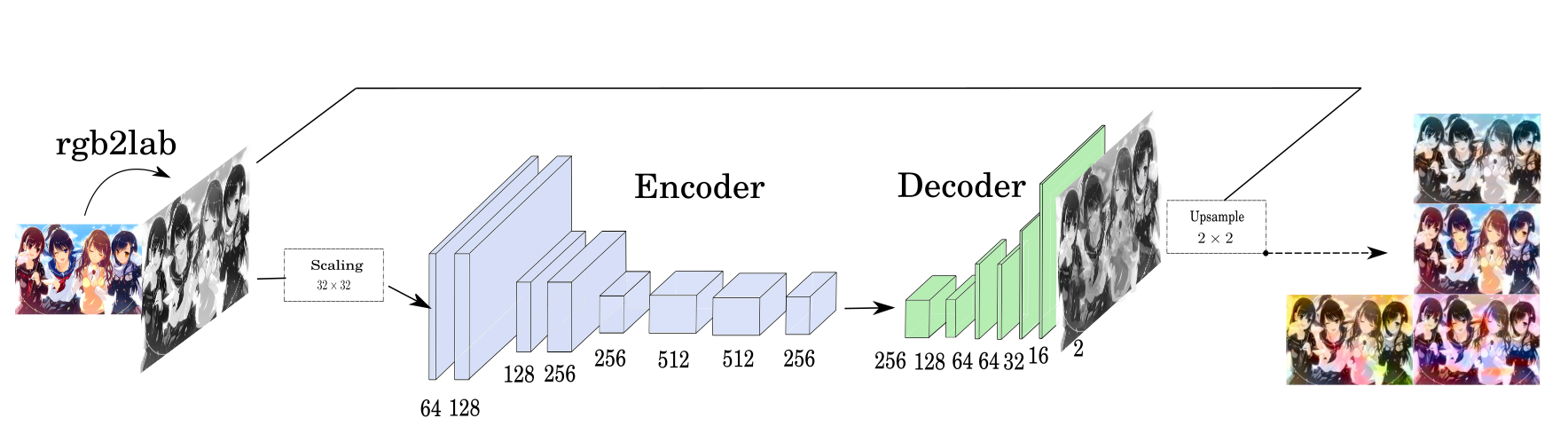}
\caption{An overview of the Encoder-Decoder image colorization model architecture.}
\label{2}
\end{figure}

\vspace{-9mm}
\section{Experiments Setup and Results}
\vspace{-2mm}
\subsection{Datasets and Model Architectures}
\vspace{-1mm}
We empirically evaluate our proposed Color-$S^{4}L$ model's performance on three widely adopted semi-supervised image classification datasets: Street View House Numbers (SVHN) \cite{42}, CIFAR-10 \cite{40}, and CIFAR-100 \cite{40}.
To make further comparisons and analysis of diverse network architectures, we first conduct experiments with 6 high-quality and widely-used network architectures on SESEMI which doesn't contain the image colorization task: (i) the 13-layer max-pooling ConvNet \cite{23,16}; (ii) the modern wide residual network with depth 28 and width 4 (WRN-28-4) \cite{13,47}; (iii) ResNet34 network with Shake-Shake regularization (Shake-WRN) \cite{49}; (iv) deep residual network ResNet50 \cite{48}; (v) 13-layer max-pooling Network-in-Network (NIN) \cite{26}; (vi) VGG16 \cite{50}, then apply the most effective architectures on Color-$S^{4}L$ embedded with the colorization task.
\vspace{-1mm}
\subsection{Color-$S^{4}L$ model Results and Analysis.} \vspace{-2mm}
Based on further research of SESEMI model\footnote{More details shown in our whole research report.}\cite{23}, we investigate our novel Color-$S^{4}L$ algorithm’s performance and analyse the \textbf{Test Classification Error Rate} separately on CIFAR-10, SVHN and CIFAR-100 datasets compared with previous supervised and semi-supervised approaches. Reviewing the research results on model architectures, we especially choose 3 outstanding CNN trunks: \textbf{ConvNet, WRN} and \textbf{Shake-WRN} to be applied with Color-$S^{4}L$ algorithm, except that we train the inference model on test dataset with 128 batch size for 30 training epochs which are significantly fewer than that utilized in \cite{22}, meaning our semi-supervised training is much efficient with less training time and computation resource. More importantly, our Color-$S^{4}L$ approach could produce image classification results competitive with, and in some cases exceeding prior state-of-the-art methods like Mixup \cite{18}, VAT SSL \cite{7} and Mean Teacher \cite{10}. 
\vspace{-5mm}
\renewcommand{\arraystretch}{1.7} %控制行高
\begin{table}
\resizebox{278pt}{56pt}{
\begin{tabular}{ccccc} 
\hline \
Method(CIFAR-10) & 1000L & 2000L & 4000L & 50000L \\
\hline
 Supervised \cite{10} & 46.43$\pm$1.21 & 33.94$\pm$0.73 & 20.66$\pm$0.57 &	5.82$\pm$0.15 \\
 Mixup \cite{18} & 36.48$\pm$0.15 & 26.24$\pm$0.46 & 19.67$\pm$0.16 & ------ \\
 Manifold Mixup \cite{18} & 34.58$\pm$0.37 & 25.12$\pm$0.52 & 18.59$\pm$0.18 & ------ \\
 SESEMI ASL(ConvNet) \cite{23} & 29.44$\pm$0.24 & 21.53$\pm$0.18 & 16.15$\pm$0.12 &  4.70$\pm$0.11\\
 \hline
 VAT SSL \cite{7}	& ------ & ------ &	\textbf{11.36$\pm$0.34} & 5.81$\pm$0.02 \\
 II Model SSL \cite{16} & ------ &	------	& 12.36$\pm$0.31 & 5.56$\pm$0.10 \\
 Mean Teacher SSL \cite{10} & 21.55$\pm$1.48 &	16.73$\pm$0.31 & \bf{12.31$\pm$0.28} & 5.94$\pm$0.15 \\
 Color-$S^{4}L$(ConvNet)	& \bf{20.45$\pm$0.34} & \bf{16.22$\pm$0.20} & 13.05$\pm$0.25 & \bf{5.07$\pm$0.06} \\
 SESEMI(ConvNet) & \bf{18.45$\pm$0.26} & \bf{15.84$\pm$0.40} & \bf{12.95$\pm$0.34} & \bf{5.21$\pm$0.24} \\
 \hline
\end{tabular}}
\caption{Color-$S^{4}L$: CIFAR-10, '1000L' means 1000 labels in the supervised branch.}
\label{tab:tb1}
\end{table} \vspace{-5mm} \\ 
\textbf{CIFAR-10}: \textbf{Table 1} presents the comparisons between Color-$S^{4}L$ model with other state-of-the-art methods based on consistency baselines. To comprehensively investigate each CIFAR-10 \& CIFAR-100 S-epochs image colorization model effects within Color-$S^{4}L$, we have conducted extensive experiments with diverse image colorization models including CIFAR-10-\{100,200,300\}-color \& CIFAR-100-100-color (where means 'Dataset-training epochs(100/200/300) colorization H5 Keras model'). Following our methodology, we find that ConvNet outperforms WRN and Shake-WRN on CIFAR-10 no matter with any colorization model and it specially accompanies best with the CIFAR-10-300-color model. Thus, we just show the best results from ConvNet model with the
CIFAR-10-300-color model in \textbf{Table 1}. Whereas, WRN represents unstable performance with most colorization models so that it can’t obtain trusty results like Shake-WRN and ConvNet, which means sometimes image colorization technique may not be complementary to the certain model architecture in Color-$S^{4}L$.

\textbf{SVHN}: Experiments on SVHN denotes a different story. The optimal performance we achieve comes from the CIFAR-10-100-color model with 30 Color-$S^{4}L$ training epochs. From \textbf{Table 2}, our newly-used Shake-WRN architecture surpasses other network architectures both in SESEMI and Color-$S^{4}L$ models. Also, we are the first to explore and excavate the good performance of Shake-WRN architecture in our novel $S^{4}L$ framework. Whereas, Color-$S^{4}L$ model can not achieve satisfactory results while compared against Mean Teacher \cite{10} which averages model weights instead of label predictions. Furthermore, as the number of labels in the supervised branch increase, Color-$S^{4}L$ with ConvNet architecture even outperforms some optimal benchmarks (e.g. SESEMI (Shake-WRN), Color-$S^{4}L$ (Shake-WRN)) with fewer labeled SVHN\footnote{\url{http://ufldl.stanford.edu/housenumbers/}} images.
\vspace{-5mm}
\begin{table*}
\begin{floatrow}
\capbtabbox{
\resizebox{168pt}{54pt}{
 \begin{tabular}{ccccc}
 \hline
 Method(SVHN) &	250L & 500L & 1000L & 73257L \\
 \hline
 Supervised \cite{10} & 27.77$\pm$3.18 & 16.88$\pm$1.30 & 12.32$\pm$0.95 &	2.75$\pm$0.10 \\
 Mixup \cite{18} & 33.73$\pm$1.79 & 21.08$\pm$0.61 & 13.70$\pm$0.47 &	------\\
 Manifold Mixup \cite{18} & 31.75$\pm$1.39 & 20.57$\pm$0.63 & 13.07$\pm$0.53	& ------\\
 SESEMI ASL(ConvNet)[23] & 23.60$\pm$1.38 & 15.45$\pm$0.79 &	10.32$\pm$0.16 & 2.26$\pm$0.07\\
 \hline
 $\prod$ Model SSL \cite{16} & ------ & 6.65$\pm$0.53 & 4.82$\pm$0.17 &	2.54$\pm$0.04 \\
 Mean Teacher SSL \cite{10} &	\bf{4.35$\pm$0.50} & \bf{4.18$\pm$0.27} & \bf{3.95$\pm$0.19} & 2.50$\pm$0.05 \\
 Color-$S^{4}L$ (ConvNet) &	17.97$\pm$0.72 & 12.92$\pm$1.26 &   	\bf{5.05$\pm$0.25} & \bf{2.57$\pm$0.06}\\
 SESEMI(ConvNet) &	16.11$\pm$1.38 & 8.65$\pm$0.18 & 5.59$\pm$0.34 &	\bf{2.26$\pm$0.07} \\
 Color-$S^{4}L$(Shake-WRN) & \bf{8.81$\pm$0.51} & \bf{6.37$\pm$0.26} &	\bf{5.13$\pm$0.12} & 3.31$\pm$0.05\\
 SESEMI (Shake-WRN) & \bf{10.52$\pm$1.36} & \bf{7.23$\pm$0.24} & 5.68$\pm$0.23 &	\bf{2.34$\pm$0.08} \\
 \hline
 \end{tabular}}
}{
 \caption{Color-$S^{4}L$: SVHN}
 \label{tab:tb3}
}
\capbtabbox{
\resizebox{148pt}{54pt}{
 \begin{tabular}{ccc}
 \hline
 Method(CIFAR-100) & 20000L & 50000L \\
 \hline
 Supervised \cite{10} & 42.83$\pm$0.24 & 26.42$\pm$0.17\\
 SESEMI ASL(ConvNet) \cite{23} & 38.62$\pm$0.31 & 22.49$\pm$0.15 \\
 ImageNet-32 Fine-tuned & \bf{30.48$\pm$0.27} & \bf{22.22$\pm$0.25} \\
 \hline
 $\prod$ Model SSL \cite{16} & 36.18$\pm$0.32 & 26.32$\pm$0.04 \\
 TempEns SSL \cite{16} & {35.65$\pm$0.41} & 26.30$\pm$0.15  \\
 Color-$S^{4}L$(ConvNet) & \bf{33.59$\pm$0.30} & \bf{24.80$\pm$0.62} \\
 SESEMI(ConvNet) & \bf{34.09$\pm$1.54} & \bf{25.32$\pm$0.39} \\
 SESEMI(WRN) & \bf{34.69$\pm$0.10} & \bf{24.83$\pm$0.12} \\
 \hline
 \end{tabular}}
}{
 \caption{Color-$S^{4}L$: CIFAR-100}
 \label{tab:tb4}
}
\end{floatrow}
\end{table*} \vspace{-4mm}

\textbf{CIFAR-100}: Semi-supervised image classification on CIFAR-100 (100 classes) seems much more challenging than CIFAR-10 and SVHN. Reviewing the image colorization section, we decided to embed CIFAR-100-100-color into CIFAR-100 Color-$S^{4}L$ since the model learned the color information from the CIFAR-100 data itself and such pretext task seems more available to extract image features and generate reasonable proxy labels. Similar to CIFAR-10, ConvNet is the best CNN trunk on CIFAR-100 dataset and we obtained competitive performance for 30 training epochs in shorter time, even our Color-$S^{4}L$ model could achieve slightly lower error rates than the retrained SESEMI model without integrating the novel image colorization self-supervised task (\textbf{Table 3}). \vspace{-3mm}

\section{Conclusions} \vspace{-2mm}
We proposed a novel Color-$S^{4}L$ model which combines multiple self-supervised pretext tasks with the normal semi-supervised learning framework. Additionally, we embedded our self-trained effective image colorization model into the SSL pipeline to establish a new supervision along with image rotation and geometric transformation. Furthermore, we explored 6 CNN architectures’ performance both in SESEMI \cite{23} and Color-$S^{4}L$ model, even discovered our first-applied Shake-WRN neural network surpasses other trunks on SVHN datasets. In sum, we dived into the interesting research on the integration of the quickly-advancing self-supervised learning and semi-supervised learning, even provided competitive or best results from our Color-$S^{4}L$ model in comparison to previous semi-supervised learning state-of-the-art methods. In the future research, we may explore more possibilities of utilizing self-supervised representation learning for other learning paradigms like few-shot learning and reinforcement learning.

\clearpage
\bibliographystyle{splncs04}
\bibliography{egbib}
\end{document}